\documentclass[runningheads]{llncs}

\usepackage{eccv}

\usepackage{eccvabbrv}

\usepackage{graphicx}
\usepackage{booktabs}

\usepackage[accsupp]{axessibility}  

\usepackage{hyperref}
\hypersetup{pagebackref=true,breaklinks=true,letterpaper=true,colorlinks,bookmarks=false,citecolor=eccvblue,linkcolor=red}

\usepackage{orcidlink}
\usepackage{makecell}
\usepackage{adjustbox}
\usepackage{booktabs}
\usepackage{subcaption}
\usepackage{multirow}
\usepackage{overpic}
\usepackage{amssymb}
\usepackage{pifont}
\usepackage{wrapfig}

\newcommand{\Paragraph}[1]{\vspace{0.6mm} \noindent \textbf{#1} \hspace{0mm}}
\newcommand{\Section}[1]{\vspace{-1mm} \section{#1} \vspace{-1mm}}
\newcommand{\SubSection}[1]{\vspace{-1mm} \subsection{#1} \vspace{-1mm}}

\newcommand{\xmark}{\ding{55}}%
\definecolor{ao(english)}{rgb}{0.0, 0.5, 0.0}
\definecolor{brightmaroon}{rgb}{0.76, 0.13, 0.28}

\begin{document}

\title{Common Sense Reasoning for Deepfake Detection} 

\titlerunning{Common Sense Reasoning for Deepfake Detection}

\author{Yue Zhang\thanks{This work was completed during an internship at Reality Defender Inc.}\inst{1,2}\and
Ben Colman\inst{2} \and
Xiao Guo\inst{1} \and
Ali Shahriyari \inst{2}\and
Gaurav Bharaj \inst{2}
}
\authorrunning{Y. Zhang et al.}

\institute{Michigan State University
\and
Reality Defender Inc\\
\email{\{zhan1624, guoxia11\}@msu.edu}}

\maketitle

\begin{abstract}
State-of-the-art deepfake detection approaches rely on image-based features extracted via neural networks. While these approaches trained in a supervised manner extract likely fake features, they may fall short in representing unnatural `non-physical' semantic facial attributes -- blurry hairlines, double eyebrows, rigid eye pupils, or unnatural skin shading. 
However, such facial attributes are easily perceived by humans and used to discern the authenticity of an image based on human \textit{common sense}.
Furthermore, image-based feature extraction methods that provide visual explanations via saliency maps can be hard to interpret for humans. To address these challenges, we frame deepfake detection as a Deepfake Detection VQA (DD-VQA) task and model human intuition by providing textual explanations that describe common sense reasons for labeling an image as real or fake. We introduce a new annotated dataset and propose a Vision and Language Transformer-based framework for the DD-VQA task. We also incorporate text and image-aware feature alignment formulation to enhance multi-modal representation learning. As a result, we improve upon existing deepfake detection models by integrating our learned vision representations, which reason over common sense knowledge from the DD-VQA task.
We provide extensive empirical results demonstrating that our method enhances detection performance, generalization ability, and language-based interpretability in the deepfake detection task.
Our dataset is available at \href{https://github.com/Reality-Defender/Research-DD-VQA}{https://github.com/Reality-Defender/Research-DD-VQA}.
\end{abstract}
\Section{Introduction}
\label{sec:intro}
\begin{figure}
    \centering
    \includegraphics[width=1\linewidth]{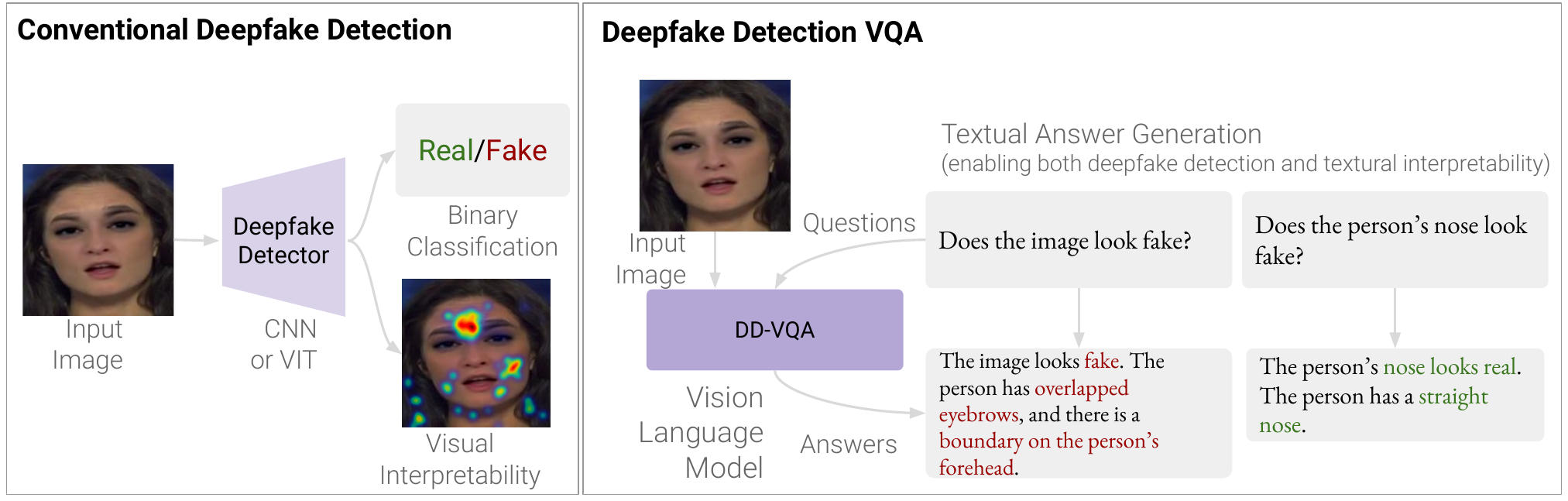}
    \caption{Illustration of the Deepfake Detection VQA (DD-VQA) Task. Conventional methods categorize deepfake detection as a binary classification task. However, we extend the task to a multi-modal task, enabling the generation of real/fake answers and corresponding explanations in response to a given question. 
    }
    \label{fig:intro}
\end{figure}
The rise of generative methods~\cite{goodfellow2020generative, karras2020analyzing, ho2020denoising, gupta2023visual} enables new capabilities to create and manipulate images. While these advances empower human creativity and enable numerous AI-for-good applications~\cite{ramesh2022hierarchical, saharia2022photorealistic}, they can also be used to create and spread misinformation, potentially leading to social problems and security threats~\cite {tolosana2020deepfakes, turton2020deepfakes, vaccari2020deepfakes}. As a result, with the increasing prevalence of generative media (deepfakes), a growing number of advanced deepfake detection algorithms~\cite{bai2023aunet, ricker2022towards, guo2023controllable} are being developed to discern media authenticity and mitigate such serious concerns.\looseness=-1

Previous deepfake detection methods primarily function as binary classifiers, including approaches such as convolution neural networks (CNNs)~\cite{shiohara2022detecting}, self-blending techniques~\cite{shiohara2022detecting} and diffusion model detection~\cite{ricker2022towards}. These methods enhance the model's interpretability via saliency maps based on visual features~\cite{selvaraju2017grad, draelos2020use, dosovitskiy2020image}. However, providing detailed explanations for the underlying reasons of authenticity or fakeness, especially in the form of explicit text explanations, remains an area with limited exploration. In fact, answering the question of ``Why the image is a deepfake?'' is a greater challenge than ``Whether the image is a deepfake?''. The former requires human common sense reasoning that is not explicit in images. We acknowledge that certain deepfake models can now generate images that effectively deceive individuals~\cite{mirsky2021creation}; however, we draw inspiration from the concept of the \textit{uncanny valley}~\cite{geller2008overcoming}, where humans occasionally experience discomfort (eeriness) for hallucinated human images that appear almost, but not quite \textit{human-like}. Therefore, it is crucial for deepfake detection methods to possess human common sense reasoning ability, enabling them to provide an explicit rationale justifying ``what's wrong'' in the image. Such ability requires the model to reference specific image regions and draw upon shared background knowledge about the typical appearance of an authentic face. However, the current deepfake detection methods lack this essential ability.

To address the above challenge, we propose a novel \textbf{D}eepfake \textbf{D}etection \textbf{V}isual \textbf{Q}uestion \textbf{A}nswer (DD-VQA) task to 
extend the deepfake detection from a binary classification task to a generative visual question-answering task (see Fig.~\ref{fig:intro}). In this task, the objective is to generate answers based on questions and images, where the answers are not limited to providing deepfake detection results but also to describe the corresponding textual explanations grounded in common sense knowledge. The common sense knowledge in our work is expressed in rich natural language, 
like the ``\textit{overlapped eyebrows}''. Our proposed DD-VQA task aims to improve deepfake detection models' common sense reasoning ability, which is crucial as the models are encouraged to focus on the cognition-perception of authenticity or fakeness, surpassing the conventional emphasis on recognition-level features in the image.

\begin{figure*}[t]
    \centering
\includegraphics[width=1.0\linewidth, height=1.5in]{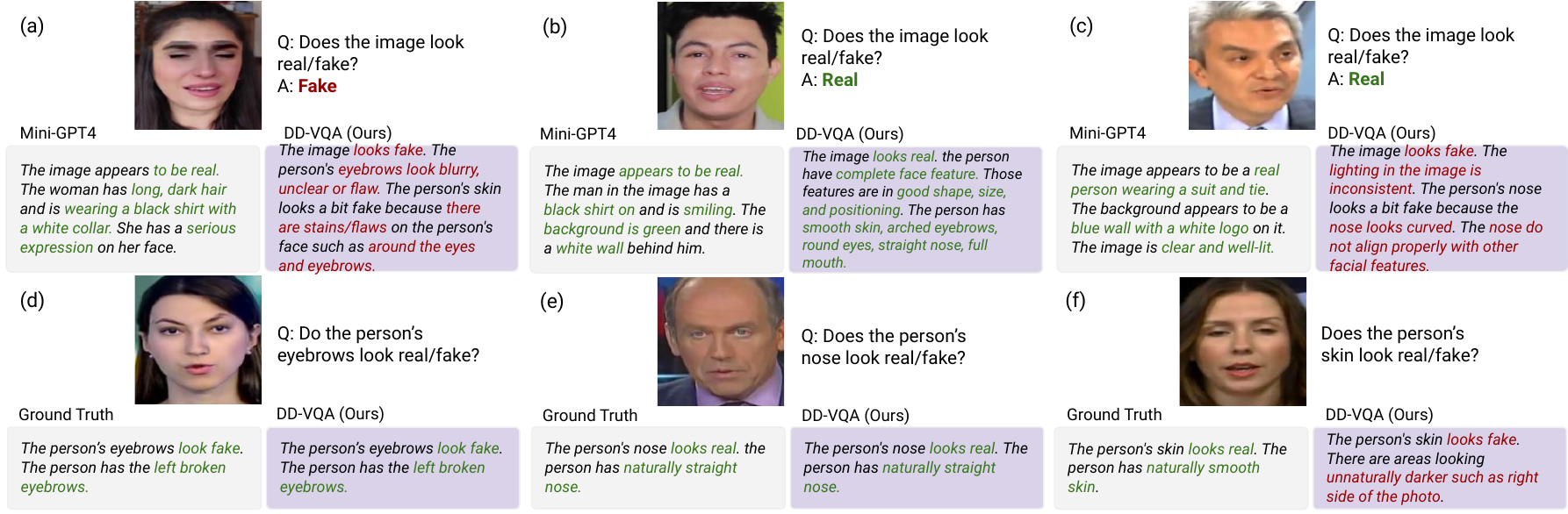}
    \caption{Qualitative Examples. 
    The first row shows \textit{MiniGPT-4~\cite{zhu2023minigpt}  vs DD-VQA (Ours)}, where (a) and (b) are successful cases, and (c) is the failure case. 
    The second row shows \textit{Ground-truth vs DD-VQA (Ours)} for fine-grained questions, where (d) and (e) are successful cases, while (f) is the failure case. The \textcolor{ao(english)}{green} and \textcolor{brightmaroon}{red} texts are the real-related and the fake-related texts, respectively.}
    \label{fig:qualitative example}
\end{figure*}

To support our proposed DD-VQA task, we \textbf{first} introduce a novel dataset, named DD-VQA dataset, including triplets of images, questions and answers. We design general and fine-grained questions for each image to inquire about the authenticity of the entire image and facial components. The annotators' answers include real/fake decisions and corresponding textual reasons.
DD-VQA task is more challenging compared to conventional binary deepfake detection task since, besides understanding the question and image, the model needs to (1) determine the authenticity of the individual facial component based on the questions asked and (2) learn common sense knowledge to generate reasonable textual explanations. 
We observe that the prevailing large VLMs~\cite{li2022blip, alayrac2022flamingo, zhu2023minigpt} encounter limitations on the DD-VQA task (see Fig.~\ref{fig:qualitative example}). Such pre-trained VL models tend to provide generic descriptions of facial features and often fall short when distinguishing image authenticity while offering reasonable explanations. \textbf{Then}, we train a VL model with the DD-VQA dataset as our proposed benchmark. We also introduce text and image contrastive losses to strengthen the model's representation learning, helping capture distinct features that differentiate between fake and real images across various modalities. We filter positive and negative images/answers based on textural answers. 
\textbf{Finally}, we integrate our cross-modal-learned visual representation into the existing deepfake detection models. 
We enhance vision representations of downstream deepfake detection with our vision representations trained on the DD-VQA datasets, improving the model's detection performance and generalization ability.
To summarize, our contributions are:

$\diamond$ We introduce a novel DD-VQA task and the corresponding dataset, enabling the generation of detection decisions and textual explanations. This task helps the deepfake detection models obtain common sense knowledge related to the image's authenticity and fakeness.

$\diamond$ We provide a multi-modal Transformer model as the benchmark and enhance representation learning through a novel text and image contrastive learning formulation. Our design helps the model reason over both textural justifications for its detection decision and the referred image region.

$\diamond$ We further employ our learned multi-modal representations in the downstream deepfake detection models to improve their detection performance and generalization ability.

$\diamond$ We evaluate the performance of DD-VQA in both aspects of deepfake detection and text generation. We also provide a comprehensive analysis to show that incorporating textual explanation can improve detection performance, generalization ability, and interpretability of the deepfake detection model.

\Section{Related Works}
\label{sec:rw}
\Paragraph{Deepfake Detection.} Deep learning methods are the dominant approaches for the deepfake detection task. The traditional CNN-based methods such as Xception~\cite{chollet2017xception}, EfficentNet~\cite{tan2019efficientnet}, and HiFi-Net~\cite{guo2023hierarchical} have achieved satisfying results in intra-dataset. 
To improve the generalization ability, Face X-ray~\cite{li2020face} identifies boundary inconsistencies as the general forgery cue. 
Some recent works explore Multi-modal models in deepfake detection task~\cite{yang2023avoid,haliassos2022leveraging}. 
However, there is very limited research integrating natural language into deepfake datasets or deepfake detection models.
VLFFD~\cite{sun2023towards} proposes a visual-linguistic paradigm to use language as supervision, 
but their text information is automatically generated and only focuses on aspects like manipulation region, type, and method. 
In contrast, our VQA dataset offers free-form textual explanations regarding the authenticity of the image based on human common-sense knowledge.


\Paragraph{Interpretable Deepfake Detection Models.}
The approaches used to interpret deepfake detection models mainly align with the methods used to explain neural network classifiers.
The prominent approach uses gradient-based methods~\cite{simonyan2014visualising, selvaraju2017grad, sundararajan2017axiomatic} to visualize the highlight regions for the prediction.
Another line of research attempts to build an interpretable network by model design~\cite{khalid2023dfgnn,trinh2021interpretable} or generating manipulation operations~\cite{shao2022detecting}; for instance, DFGNN~\cite{khalid2023dfgnn} applies interpretable GNN to deepfake detection tasks. 
While these methods have been used to enhance the model's interpretability, describing reasons for the determination in natural language has yet to be explored extensively.


%

\Paragraph{Vision-Language Learning.}
Multi-modal learning, especially Vision-Language (VL) learning, has gained significant attention within the AI community. 
Recently, an increasing number of large VL pre-training models have emerged, such as BLIP~\cite{li2022blip}, Flamingo~\cite{alayrac2022flamingo}, and MiniGPT4~\cite{zhu2023minigpt}.
These models are all based on the Transformer architecture and have been trained on various VL datasets and tasks~\cite{antol2015vqa, anderson2018vision, zhang2023vln, zhang2022lovis, zhang2024navhint,vinyals2016show,guo-multi-domain,agrawal2019nocaps}. 
Despite their popularity, relatively little research has been dedicated to examining these models' performance in deepfake detection.

\Paragraph{Visual Question Answering} has been one of the most popular research topics. The task requires reasoning ability on visual images and textual questions to predict the correct answer based on specific knowledge.
A large-scale VQA dataset with free-form questions created by humans is first proposed in~\cite{antol2015vqa}. 
Vision Commonsense~(VCR)~\cite{zellers2019recognition} is a VQA dataset to provide a rationale to justify its answers based on the details of the movie scene and background knowledge about how the world works.
We are pioneering in extending deepfake detection into the research area of the VQA task. 


\Section{Our Method}
\begin{table*}[t]
\scriptsize
    \centering
    \begin{adjustbox}{width=\linewidth,center}
    \begin{tabular}{p{6pt}|p{68pt}|p{190pt}|p{180pt}}
        \hline
         \# &\textbf{Facial Features} & \textbf{Fake Features} & \textbf{Real Features} \\
         \hline
        1 & Eyebrows & overlapping, broken, blurred, etc. & arched, straight, thick, thin, bushy, sparse, etc.  \\
        2 & Skin & boundaries, stain, flaws, inconsistent color, etc. & smooth, illuminated, wrinkled, even, etc. \\
        3 & Eyes &  overly large, small, blurred, too rigid, etc. & round, oval, deep, large, small, sparking, etc. \\
        4 & Nose & unnaturally curved, lack of details, no nose, etc. & straight, pointed, broad, etc. \\
        5 & Mouth/Teeth/Chin & overly large, small mouth, unnatural color for mouth, teeth, overly pointed, square chin, etc. & full, thin, pouty mouth, white/aligned/misaligned teeth, overly squared/ pointed chins, etc. \\
        6 & Others & mismatched bangs/fringe/mustache/beard, blurry eyeglasses frame, inconsistent/unrealistic shadow, etc. & complete face feature, proper hair/bangs/hairstyle, matched mustache/beard, etc.\\
    \hline
    \end{tabular}
    \end{adjustbox}
    \caption {Annotation vocabulary for the reasons of authenticity or fakeness based on common sense knowledge.}
    \label{tab:vocabulary}
\end{table*}
We present our method in the following sections. In Sec~\ref{DD-VQA Dataset}, we introduce our methods of constructing the DD-VQA dataset. We introduce our proposed model architecture and designed learning objectives for the DD-VQA task in Sec~\ref{sec:model} and Sec~\ref{DD-VQA objectives}, respectively. In Sec.~\ref{enhancement}, we introduce our method utilizing the learned multi-modal representation to enhance deepfake detection models.

\SubSection{DD-VQA Dataset Collection}
\label{DD-VQA Dataset} 
Deepfake Detection VQA task is to generate answers given an image and a question to discern facial authenticity. 
To the best of our knowledge, we are pioneering research generating textual explanations for the deepfake detection task. Thus, no existing dataset can be directly used for this purpose. We present our methodology for constructing the DD-VQA dataset.

The manipulated images are collected from the FaceForensics++ dataset (FF++)~\cite{rossler2019faceforensics++}. The FF++ dataset, widely used in deepfake research, contains $5,000$ videos. Among these, $1,000$ videos are real, while $4,000$ videos are fake, employing four different manipulation methods: \textit{Deepfakes},  \textit{Face2Face}, \textit{FaceSwap}, and \textit{NeuralTextures},
While we acknowledge the existence of deepfake datasets that are even more effective at deceiving individuals~\cite{mirsky2021creation}, our goal is to explore the use of text to explain obvious deepfake indicators. For this purpose, we use the FF++ dataset, which is relatively easier to identify and describe manipulation regions. The FF++ dataset solely provides video frames, and our primary focus is to assess the authenticity of the human faces in the image. So, in our initial step, we extract one frame from each video and crop the human face from the frame. The image is sized at $480\times 480$ pixels with complete human faces. 
After collecting images, we gather the corresponding question-answer pairs based on our designed annotation schemes, as introduced below.

\Paragraph{Annotation Scheme}
We utilize Amazon Mechanical Turk~\footnote{\url{https://www.mturk.com/}} as our annotation platform.
During the annotation process, we observe cases where the entire face is identified as fake, yet specific facial components appear authentic. For instance, in the example of Fig.~\ref{fig:intro}, the person's \textit{overlapped eyebrows} is an obvious indicator of fakeness, whereas the person's \textit{naturally straight nose} appears relatively realistic. 
In such cases, we frame our task as a VQA task, tasking annotators to answer specific questions. We design two types of questions for each image: \textbf{\textit{general questions}} and \textbf{\textit{fine-grained questions}}.
The general question requires an evaluation of overall facial authenticity, while the fine-grained question is to assess the authenticity of facial components, mainly including \textit{skin}, \textit{eyebrows}, \textit{eyes}, \textit{nose}, and \textit{mouth}. 
The answers to both questions are expected to be binary yes-or-no answers and detailed factors based on common sense knowledge. 

\begin{figure*}[t]
    \centering \includegraphics[width=\linewidth, height=1.8in]{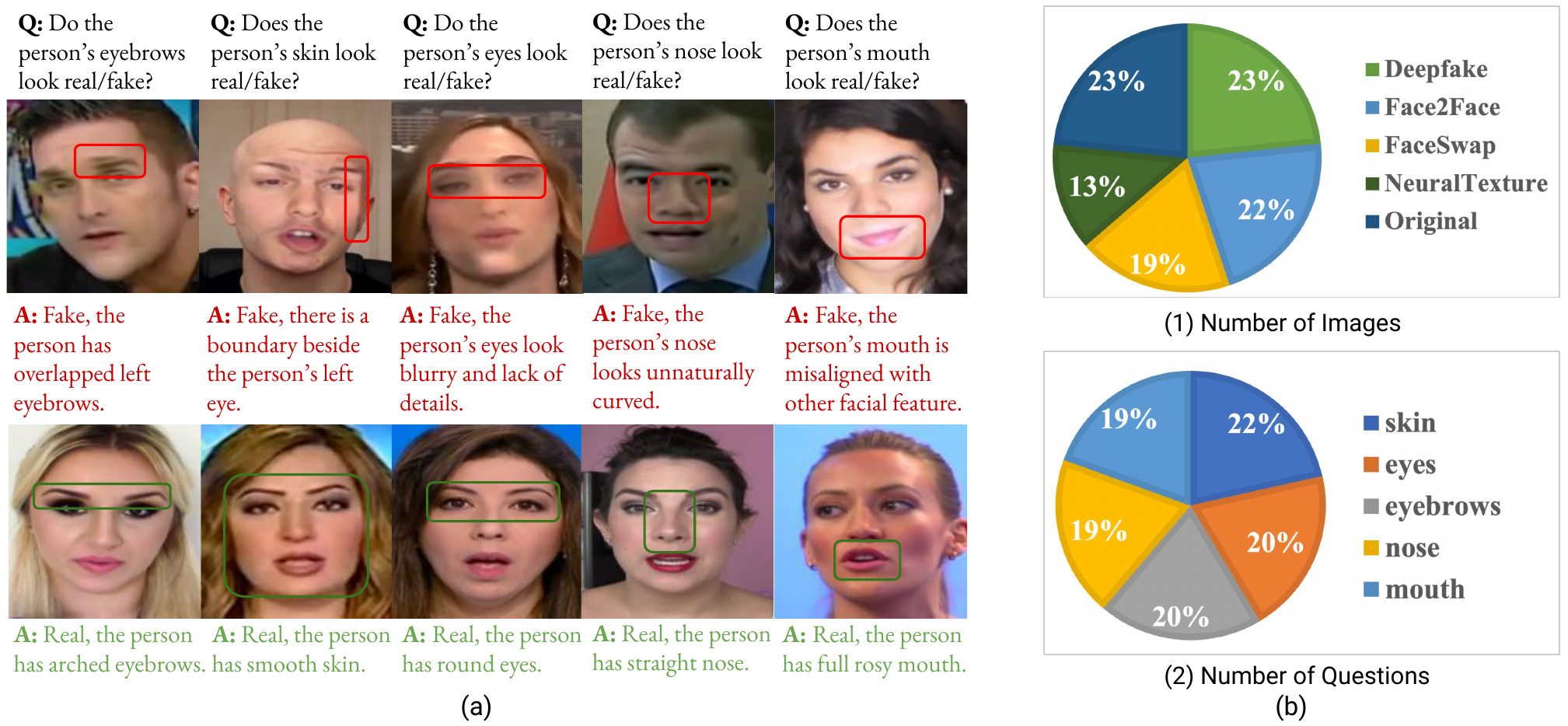}
    \caption{DD-VQA Dataset. (a) Examples of fine-grained question-answer pairs. (b) Statistics of the DD-VQA dataset.
    }
    \label{fig:DD-VQA Dataset}
\end{figure*}

\Paragraph{Annotation Challenges} After collecting the initial batch of annotations, we observe that annotators struggle to provide detailed descriptions of the characteristics of authenticity or fakeness.
For example, many annotations contain only detection results, leaving the explanation part blank. Even when explanations are provided, annotators tend to use very general terms, such as ``\textit{unrealistic skin}'' or ``\textit{unnatural eyebrow}''. To address this issue, we design a set of predefined reasons based on human common-sense knowledge. We list some descriptive expressions to specify the reasons for the authenticity and fakeness of different facial components in Tab.~\ref{tab:vocabulary}. These expressions are formulated as multiple-choice lists for annotators to select when responding to each question. Additionally, annotators maintain the flexibility to provide additional explanations, contributing to refining our answer lists. 
Beyond appearance-related reasons, we also include expression-related reasons, such as \textit{``furrowed eyebrows''}, \textit{``rigid eyes''}, and \textit{``rigid mouth''}, which are also crucial to deepfake detection. 
Please refer to the supplementary for more details about various fake features and annotations.
\Paragraph{Statistics of the DD-VQA Dataset}
Given the variation in people's perceptions of authenticity, we gather annotations from three annotators for each image. In the detection aspect of the answer (fake/real), we adopt the majority choice, wherein we gather annotations when at least two annotators agree. We will keep all the provided explanations in the answer if an agreement is reached. As a result, up to three answers are collected for each question. Annotators can skip the question if they cannot answer, resulting in various numbers of question-answer pairs for each image.
We averagely collect approximately $3$ to $6$ question-answer pairs for each image. Then, we remove low-quality annotations based on the following criteria:
\textbf{1)} Absence of answers to all questions for an image.
\textbf{2)} Conflicting annotations where annotators select both ``real'' and ``fake'' labels for the same question, which means the annotators are unsure about their decisions.
\textbf{3)} Annotations differ from the ground-truth detection labels. We value how people perceive the authenticity of an image. If an image successfully fools the annotators, their answers to the question may not be reliable.

\begin{figure*}[t]
  \centering
\includegraphics[width=1.\linewidth, height=1.6in]{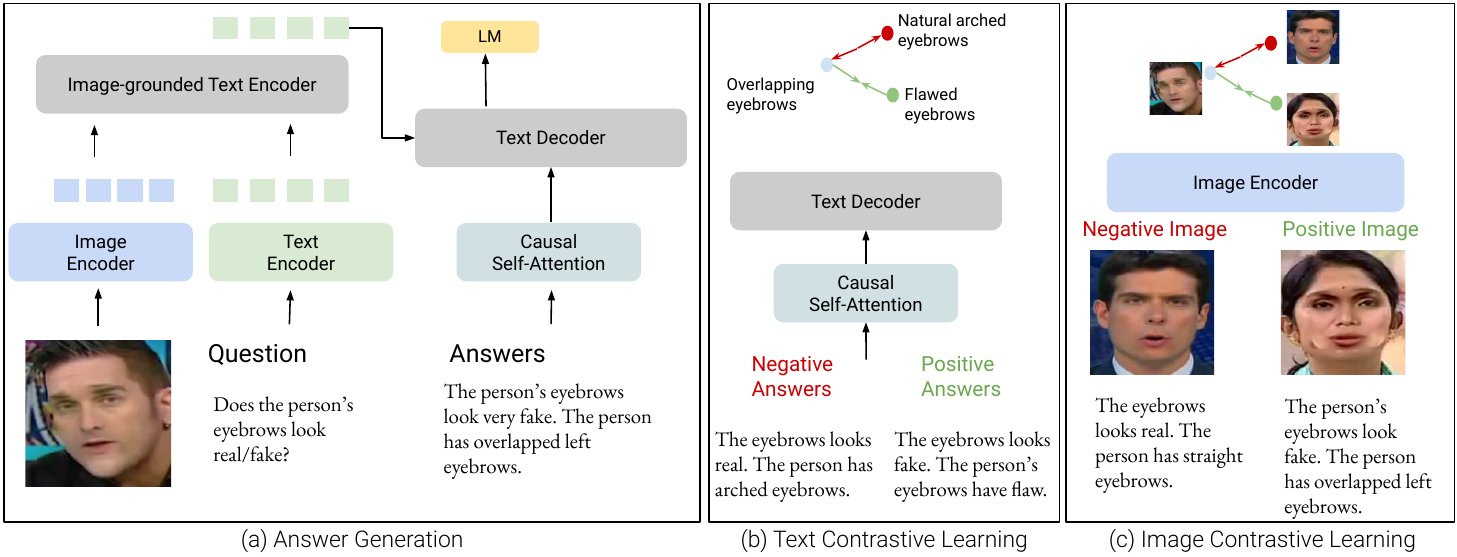}
    \caption{DD-VQA Model Architecture. Our model takes the image and question as input and generates textual answers auto-regressively, as shown in (a). To enhance representation learning,  we explore two contrastive losses. In (b), we gather negative and positive answers to optimize the text encoder and decoder. In (c), we use answers to filter the negative and positive images to optimize the image encoder. }
    \label{fig:short-a}
  \label{fig:model architecture}
\end{figure*}

After collecting the annotation of answers for each question, we employ a template to post-process the annotators' choices and any additional reasons they provide. The template of the answer is ``\textit{The X looks real/fake because X looks Y}'', \textit{X} represents the entire image or any facial component, and \textit{Y} denotes the corresponding reason. In cases with multiple reasons, we use commas to combine them as the final answer. Additionally, for general questions, except for collecting the provided general reasons, we randomly select two reasons from the fine-grained answers of the same image as its complementary reasons. This helps to answer general questions more comprehensively.

In conclusion, we collect a dataset of $2,968$ images and $14,782$ question-answer pairs. We provide statistics of the DD-VQA dataset in Fig.~\ref{fig:DD-VQA Dataset}~(b) from two aspects: 1) Fig.~\ref{fig:DD-VQA Dataset}~(b)(1) shows the distribution of different manipulated methods. The distribution is relatively even while the number of images manipulated by \textit{NeuralTexture} technique are relatively smaller, indicating this technique relatively fools people more than other manipulation techniques.
2) Fig.~\ref{fig:DD-VQA Dataset}~(b)(2) shows the distribution of different facial components in those questions, indicating an even distribution among different facial components.

\SubSection{DD-VQA Model Architecture}
\label{sec:model}
Unlike conventional methods that model the deepfake detection task as a binary classifier, we propose a generative approach to model the DD-VQA task. We aim to identify the relevant image region based on the question and generate a reasonable answer based on common sense knowledge.
We adopt BLIP~\cite{li2022blip} as our backbone, a robust Transformer-based VL model pre-trained on noisy web data and bootstraps captions. BLIP is a strong backbone due to its competitive performance across various vision and language tasks, as well as its ease of training. We demonstrate our model architecture in Fig.~\ref{fig:model architecture}.


\Paragraph{Text Encoder.} 
We apply a BERT~\cite{devlin2018bert} tokenizer to split the questions into a sequence of tokens, denoted as $Q=\{\texttt{[CLS]}, q_1, q_2, \cdots, q_l,\texttt{[SEP]}\}$, where $l$ is the length of question tokens. $\texttt{[CLS]}$ and $\texttt{[SEP]}$ are the special tokens. The question is passed through a text encoder, which consists of a multi-layer self-attention block and obtains the text representations, denoted as $X_q=[x_{q1}, x_{q2}, \cdots, x_{ql}]$.

\Paragraph{Image Encoder.} We use a Vision Transformer (ViT)~\cite{dosovitskiy2020image} as our image encoder. The image is first divided into $m$ patches, and then encoded as a sequence of embedding with $\texttt{[CLS]}$ token as the global image representation. We define the obtained vision representations from ViT as $I=[i_1, i_2, \cdots, i_m]$.

\Paragraph{Image-grounded Text Encoder.} We apply cross-modal attention layers between the text representations of question $X_q$ and visual representations of image $V$ to inject visual information into the question. We obtain the attended text representation $\bar{X}_q$ as follows,
\begin{equation}
\small
    \bar{X}_q = \texttt{cross\_attn}(Q=X_q,K=I,V=I),
\end{equation}
where $Q$, $K$, and $V$ are the query, key, and value of attention, respectively.

\Paragraph{Text Decoder.} 
Similar to a question, we obtain answer tokens using a BERT tokenizer, denoted as $A=\{\texttt{[CLS]}, a_1, a_2, \cdots, a_k,\texttt{[SEP]}\}$, where $k$ is the length of tokens. We also acquire text representations of the answer through BERT. However, instead of utilizing self-attention as in the text encoder for the question, we employ causal self-attention layers to attend only to the previous tokens instead of all tokens. We represent the text representations of the answer as $X_a=[x_{a1}, x_{a2}, \cdots, x_{ak}]$. Subsequently, another cross-modal attention layer for decoding the next token is applied between the attended question representation $\bar{X_q}$ and $X_a$, as follows,
\begin{equation}
\small
    \bar{X_a} = \texttt{cross\_attn}(Q=X_a, K=\bar{X_q},V=\Bar{X_q}),
\end{equation}
where $\bar{X_a}$ is the attended answer representation given image and question, which is then fed to a Multi-layer Perceptron (MLP) to predict the answer tokens.

\SubSection{DD-VQA Learning Objectives}
\label{DD-VQA objectives}
We use three objectives to train our model. We utilize language modeling to generate answer tokens. Besides, we design text and image contrastive losses to leverage annotated textual information to filter positive and negative examples based on questions. Such contrastive learning enhances the model's multi-modal representation learning and common sense reasoning by capturing the distinctions between real and fake features.

\Paragraph{Language Modeling} aims to generate answer tokens auto-regressively, given a question and an image.  Specifically, it is a cross-entropy loss that maximizes the likelihood of the answer tokens conditioned on the previous tokens and the attended question representations, as follows:
\begin{equation}
\small
\mathcal{L}_{LM}=-\sum_{j=1}^{k} \log p_\theta(a_j |
a_1, \cdots, a_{j-1}, \bar{X}_q),
\end{equation}
where $\theta$ denotes the model's trainable parameters.
There are a maximum of three candidate answers for each question, and we compute the average loss of all answers.

\begin{wrapfigure}{r}{0.6\textwidth}
    \includegraphics[width=1\linewidth]{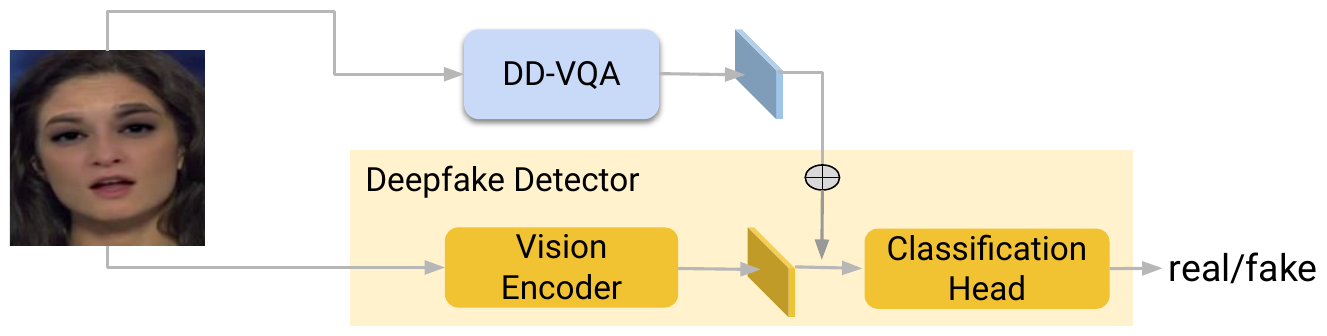}
    \caption{
    DD-VQA Enhanced Deepfake Detection. 
    We incorporate representations extracted from DD-VQA into any existing deep fake detector containing a vision encoder and classification head. 
    }
    \label{fig:fine-tune}
\end{wrapfigure}


\Paragraph{Text Contrastive Learning} aims to train the model with different answers given the same images and questions. We filter a negative and a positive answer based on ground-truth answers. The negative answers are obtained by choosing answers on the same facial component but with the opposite detection results. For instance, in Fig.~\ref{fig:model architecture}~(b), the negative answer is the description of ``\textit{real eyebrows}'' given the ground-truth ``\textit{fake eyebrows}''. The positive answer is randomly selected from the candidate answers of the current example. We input negative and positive answers to the text encoder and text decoder and use the corresponding representation of the $\texttt{[CLS]}$ token representation from the last layer of the decoder to do text contrastive learning. We denote anchor, positive and negative answer representation as $\bar{X}_{aa}$, $\bar{X}_{ap}$ and  $\bar{X}_{an}$, respectively. Then, the InfoNCE loss~\cite{oord2018representation} is used for contrastive learning as follows,
\begin{equation}
\scriptsize
\mathcal{L_T} = -log\frac{\exp(\bar{X}_{aa}\cdot \bar{X}_{ap})/\tau}
{(\exp(\bar{X}_{aa}\cdot \bar{X}_{ap}) + \exp(\bar{X}_{aa}\cdot \bar{X}_{an}) )/\tau},
\end{equation}
where $\tau$ is the temperature. The aim is to learn the attended text representation that is close to the positive answer but far apart from the negative one.

\Paragraph{Image Contrastive Learning} aims to learn the visual representation that can help the model generate correct answers. We train the model with different images given the same question and answer. We filter the positive and negative images based on the answer of the input image. For example, in Fig.~\ref{fig:model architecture}~(c), when the answer is ``\textit{overlapped eyebrows}'', the positive image is the one annotated as ``\textit{overlapped eyebrows}'' and the negative image is the one annotated as ``\textit{natural eyebrows}''. After obtaining positive and negative images, we optimize the image encoder. We use $\texttt{[CLS]}$ token representation from ViT to do image contrastive learning. We represent the anchor, positive and negative image representation as $i_{a}$, $i_{p}$, and $i_{n}$, and calculate losses as follows,
\begin{equation}
\scriptsize
\mathcal{L_I} = -log\frac{\exp(i_{a}\cdot i_{p})/\tau}
{(\exp(i_a\cdot i_{p}) + \exp(i_a\cdot i_{n}) )/\tau}
\end{equation}

We sum the three above losses as the final objective to train the model: $\mathcal{L} = \mathcal{L}_{LM} + \mathcal{L_T} + \mathcal{L_I}$.

\SubSection{DD-VQA Enhanced Deepfake Detection}
\label{enhancement}
We utilize the learned multi-modal representation from the proposed DD-VQA to augment the vision representation of the current deepfake detection model, enhancing its binary detection performance, as depicted in Fig.~\ref{fig:fine-tune}.
Moreover, the DD-VQA feature enhancement is model-agnostic, as Tab.~\ref{tab:Intra-testing Comparision} indicates.

More formally, given an image, we obtain vision representations of the DD-VQA model and deepfake detector, which are denoted as $\mathbf{F} \in \mathbb{R}^{W \times H \times C}$ and $\mathbf{F}^{\prime} \in \mathbb{R}^{W' \times H' \times C'}$, respectively. Then we can have the enhanced deepfake detector vision representations $\mathbf{F}^{en.}$ as $\mathbf{F}^{en.} = \mathbf{F}^{\prime} + \theta(\mathbf{F})$, where $\theta(*)$ represents the necessary tensor shape transformations for fusing $\mathbf{F}$ and $\mathbf{F}^{\prime}$. More details are in the supplementary.

\Section{Experiment}
\label{sec:results}
\SubSection{Experimental Setup}
\Paragraph{DD-VQA Dataset.}
Our proposed DD-VQA dataset includes $14,782$ question-answer pairs. Following the FF++ train/test ids, we partition the dataset into training and testing sets, resulting in $13,559$ question-answer pairs for training and $1,223$ for testing. The training dataset contains $2,726$ images, while the test dataset contains $242$ images.

\Paragraph{DD-VQA Evaluation Metrics.}  We evaluate the generated answer from two aspects for the DD-VQA benchmark: the performance of deepfake detection and the quality of generated explanations. We use Accuracy, Precision, Recall and F1-Score metrics to assess detection performance. 
In the DD-VQA task, we do not use the AUC metric for detection, given that our generated space is whole vocabulary text tokens of BERT rather than binary outputs. 
To assess the explanation quality, we use natural language generation metrics such as BLUE-4~\cite{papineni2002bleu}, CIDEr~\cite{vedantam2015cider}, ROUGE\_L~\cite{lin2004rouge}, METEOR~\cite{denkowski2014meteor}, and SPICE~\cite{anderson2016spice}. These scores evaluate the similarity between the generated and the annotated answer tokens.
Please see more details about evaluation metrics in the supplementary.

\Paragraph{Deepfake Detection Evaluation Metrics.} We also evaluate the effectiveness of our learned multi-modal representations on the existing deepfake detection methods. The evaluation is based on commonly used metrics for deepfake detection, including Accuracy(Acc), Area Under the Receiver Operating Characteristic Curve (AUC), and Equal Error Rate~(EER).

\Paragraph{Implementation Details.}
Our models are implemented in PyTorch~\cite{paszke2019pytorch}. 
We use BLIP-base weights as our initial pre-training weights, and the image transformer is ViT-B/16.
When fine-tuning BLIP on our DD-VQA dataset, we conduct $300$ epochs using $3$ NVIDIA RTX A6000 GPUs~($72$ hours), with a batch size of $8$ and a learning rate of $2e-5$. We use AdamW~\cite{loshchilov2017decoupled} as the optimizer with a weight decay of $0.05$. 
During inference, the max generated token is set as $50$.

\SubSection{Results on the DD-VQA dataset}
We fine-tune BLIP on the DD-VQA dataset and provide results for both deepfake detection and the corresponding explanations. Our model effectively captures the answer templates of ``\textit{The X looks real/fake. The person's X looks Y}''. Additionally, we observe an absence of cases where the detection results conflict with their corresponding explanations. In such a case, we split the generated text into sentences and extracted the tokens of  ``fake'' or ``real'' in the first sentence to assess the deepfake detection performance. We conduct an ablation to assess the impact of our proposed contrastive losses on top of the baseline. 

Table~\ref{tab:main results} shows our results. Row\#1 is the result after fine-tuning BLIP with the DD-VQA dataset. On top of it, we add text contrastive loss~(row\#2), and both deepfake detection performance and answer generation quality improve. In row\#3, we present results obtained by training the model with LM and image contrastive loss. The results show that image contrastive loss is more effective than text contrastive loss. The best result in row\#4 is achieved by training the model with all losses, resulting in an improvement of nearly 3\% in F1 and 6\% in accuracy over the BLIP baseline~(Row\#1). This result indicates our designed contrastive losses help the model distinguish real or fake features more effectively.

\begin{table*}[t]
\begin{adjustbox}{width=\linewidth,center}
\begin{tabular}{rrcccccccccc}
\hline
  \#  & Method & \multicolumn{4}{c}{DD-VQA Deepfake Detection} & \multicolumn{5}{c}{DD-VQA Answer Generation}  \\
   \cmidrule(r){3-6} \cmidrule(r){7-11} 
   && Acc $\uparrow$ & Recall $\uparrow$ & Precision $\uparrow$& F1 $\uparrow$ & BLUE-4 $\uparrow$& CIDEr $\uparrow$ & ROUGE\_{L} $\uparrow$& METEOR$\uparrow$ & SPICE$\uparrow$ \\
   \hline
    1 & BLIP~\cite{li2022blip} (baseline) & $0.8168$ & $\mathbf{0.9596}$ & $0.7861$ & $0.8642$ & $0.3569$ & $1.8177$ & $0.5664$ & $0.3301$ & $0.6658$\\
    2 &  BLIP-T (ours) & $0.8365$ &  $\underline{0.9489}$ & $0.8131$ & $0.8758$ & $0.3714$ & $ 1.8715$ &  $0.5774$ & $0.3349$ & $0.6710$ \\
    3 & BLIP-I (ours) & $\underline{0.8487}$ & $0.9448$ & $\underline{0.8298}$ & $\underline{0.8836}$ & $\underline{0.3800}$ & $\underline{1.8931}$ & $\underline{0.5882}$ & $\underline{0.3419}$  &$\underline{0.6788}$ \\
    4 & BLIP-TI (ours)  & $\mathbf{0.8749}$ & $0.9341$ & $\mathbf{0.8697}$ & $\mathbf{0.9007}$ & $\mathbf{0.4075}$ &  $\mathbf{2.0567}$ & $\mathbf{0.6085}$ & $\mathbf{0.3463}$ & $\mathbf{0.6915}$\\
\hline
\end{tabular}
\end{adjustbox}
\caption{
Experimental results of fine-tuning BLIP with the DD-VQA dataset, including both deepfake detection and answer generation performance.
Deepfake detection results in DD-VQA are based on generated texts rather than binary classification scores, so we do not provide an AUC metric. 
BLIP-T and BLIP-I represent BLIP with text and image contrastive losses, respectively, while BLIP-TI denotes BLIP with both text and image contrastive losses.
[Key: \textbf{Best}, \underline{Second Best}]
}
\label{tab:main results}
\end{table*}

\SubSection{Results on Deepfake Detection Models}
We assess the effectiveness of the learned multi-modal representations by incorporating them into existing deepfake detection models. We experiment with the following deepfake detectors: XceptionNet~\cite{rossler2019faceforensics++}, HiFi-Net~\cite{guo2023hierarchical}, SBIs~\cite{shiohara2022detecting} and RECCE~\cite{cao2022end}. 
We conduct both intra-testing and cross-testing. Specifically, we train the model on the challenging dataset of c40, a low-quality setting of FF++. We then assess the model's cross-testing performance on Celeb-DF~\cite{Celeb_DF_cvpr20} and WildDeepfake~\cite{zi2020wilddeepfake} datasets.

\begin{table}[t]
\begin{adjustbox}{width=0.95\linewidth,center}
    \begin{tabular}{r rccccccccc}
    \hline
        \multirow{3}{*}{\#} &\multirow{3}{*}{Methods} & \multirow{3}{*}{\shortstack[c]{Multi-modal\\ Enhancement}} & \multicolumn{2}{c}{Intra-testing} & \multicolumn{6}{c}{Cross-testing}\\
        \cmidrule(r){4-5} \cmidrule(r){6-11} 
        && & \multicolumn{2}{c}{FF++~(c$40$)} & \multicolumn{3}{c}{Celeb-DF} & \multicolumn{3}{c}{WildDeepfake}\\
        \cmidrule(r){4-5} \cmidrule(r){6-8} \cmidrule(r){9-11} 
        & & &ACC$\uparrow$ & AUC$\uparrow$ & ACC$\uparrow$ &AUC$\uparrow$& EER$\downarrow$ & ACC$\uparrow$ & AUC$\uparrow$& EER$\downarrow$\\
        \hline
        $1$ & XceptionNet~\cite{rossler2019faceforensics++} & \xmark & $86.86$ & $89.30$ & - & $61.80$ & $41.73$ & - & $62.72$ & $40.65$  \\
        $2$ & XceptionNet(ours) & BLIP-TI & $89.25$ & $92.24$ & $62.41$ & $64.30$ & $36.51$ & $62.52$ & $64.53$ & $39.42$ \\
        \hline
        $3$ & HiFi-Net~\cite{guo2023hierarchical} & \xmark & $89.16$ & $92.10$ & $67.20$ & $68.80$ & $36.13$ & $66.29$ & $65.22$ & $38.65$  \\
        $4$ & HiFi-Net (ours) & BLIP-TI & $\underline{91.25}$ & $95.14$ & $\underline{69.37}$ & $71.00$ & $\underline{35.70}$ & $\mathbf{69.27}$ & $\mathbf{70.03}$ & $\mathbf{35.21}$ \\
       \hline
    $5$ & SBIs~\cite{shiohara2022detecting} & \xmark & $-$ & $\underline{99.64}$ & -  & $\underline{93.18}$ & $-$ & $-$ & $-$ & $-$\\
    $6$ & SBIs~(ours) & BLIP-TI & $-$ & $\mathbf{99.67}$ & - & $\mathbf{93.98}$ & $-$ & $-$ & $-$ & $-$\\
    \hline 
     $7$ & RECCE~\cite{cao2022end} & \xmark & $91.03$ & $95.02$ & $67.96$  & $68.71$ & $35.73$ & $62.03$ & $64.31$ & $40.53$\\
    $8$ & RECCE-BLIP & BLIP & $89.22$ & $93.71$ & 
   $68.07$ & $68.36$ & $36.64$ & $65.09$ & $67.24$ & $37.49$ \\
   $9$ & RECCE (ours) & BLIP-TI & $\mathbf{92.08}$ & $95.36$ & $\mathbf{69.46}$ & $70.21$ & $\mathbf{35.63}$ & $\underline{66.57}$ & $\underline{69.46}$ & $\underline{35.90}$\\
    \hline
    \end{tabular}
    \end{adjustbox}
    \caption{The multi-modal enhanced deepfake detection performance. 
    BLIP-TI represents BLIP fine-tuned for the DD-VQA task. 
    [Key: \textbf{Best}, \underline{Second Best}]}
    \label{tab:Intra-testing Comparision}
\end{table}

As shown in Table~\ref{tab:Intra-testing Comparision},  
the improved results in row\#2, \#4,\#6, \#9 indicate that our learned multi-modal representations help improve the detection performance of all baselines in both intra-testing and cross-testing across various evaluation metrics. 
Please note that SBIs only uses real images from FF++(raw) for training. 
This experiment setup enhances SBIs's generalization ability to general manipulation types, which helps achieve better performance on FF++(c$40$) and Celeb-DF than HiFi-Net and RECCE, which are trained on both real and fake images from FF++. 
We conduct an additional experiment for the RECCE, fine-tuning the model with BLIP initial pre-trained weights instead of BLIP weights training using the DD-VQA dataset, as shown in row\#8.
We notice that utilizing BLIP initial weights already contributes to improving the generalization ability, as evidenced by the improved results in cross-testing. 
This result highlights representations learned from text modality help the deepfake detection. 
Moreover, fine-tuning BLIP with the DD-VQA dataset~(BLIP-TI) further improves another 2\% in both intra-testing and cross-testing~(row\#9). This suggests that the vision representation obtained through the DD-VQA task, which incorporates textual explanations regarding the image's authenticity based on common sense knowledge, proves beneficial for deepfake detection. 

\begin{table}[t]
    \centering
   \begin{subfigure}[t]{0.5\textwidth}
        \centering
        \includegraphics[width=0.98\linewidth, height=1in]{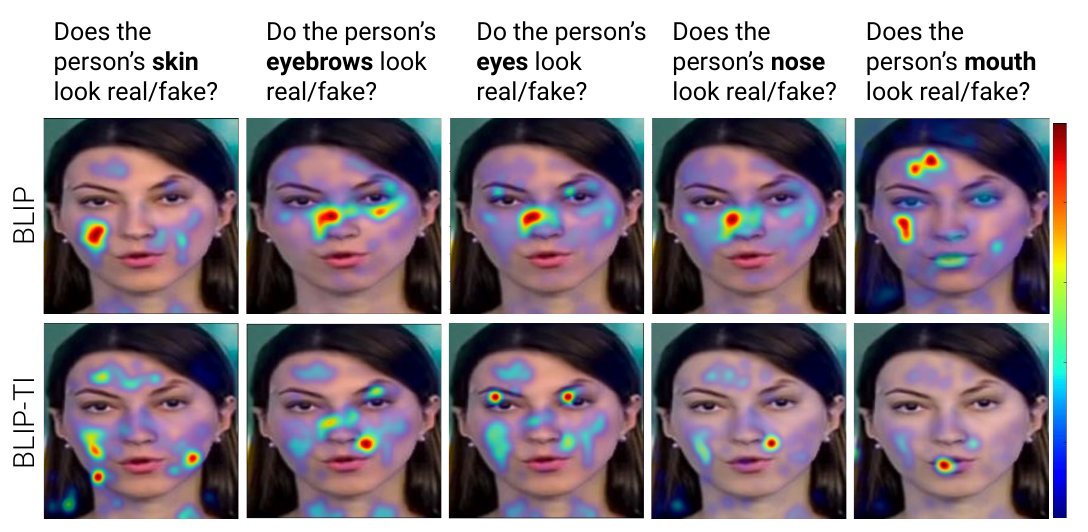}
    \end{subfigure}
    \begin{subtable}[b]{0.45\linewidth}
        \footnotesize
       \centering
    \begin{adjustbox}{width=0.98\linewidth, center}
    \begin{tabular}{r c c c c}
    \hline
        Method & Acc$\uparrow$ & Recall$\uparrow$ & Precision$\uparrow$ & F1$\uparrow$\\
        \hline
        Eyebrows & $0.8645$ & $0.9189$ & $0.8571$ & $0.8870$ \\
        Skin & $\mathbf{0.8899}$ & $\mathbf{0.9534}$ & $\mathbf{0.8786}$ & $\mathbf{0.9145}$ \\
        Eyes & $0.8750$ & $0.9224$ & $0.8770$ & $0.8992$  \\
        Nose & $0.8800$ & $0.9237$ & $0.8790$ & $0.9008$ \\
        Mouth & $0.8670$ & $0.9459$ & $0.8468$ & $0.8936$ \\
      \hline 
    \end{tabular}
    \end{adjustbox}
    \label{}
    \end{subtable}
\caption{LEFT: Attention heatmap visualization of the baseline BLIP (top) and BLIP-TI (bottom). RIGHT: Deepfake detection performance on fine-grained questions.}
   \label{tab:ablation fine}
\end{table}

\SubSection{Ablation Study}
\label{sec:ablation}
\Paragraph{Do explanations help in deepfake detection?} We conduct an experiment to train BLIP on the DD-VQA dataset, comparing the model's performance with and without corresponding explanations. In the cases without explanations, the answer template for each question is only ``\textit{The X looks real/fake.}''. The results in Tab~\ref{tab:ablation}~(left) indicate that the detection performance is higher when explanations are included, demonstrating that the rich common-sense knowledge in our designed explanations is beneficial to deepfake detection.
We also perform cross-testing of our model on SeqDeepfake dataset~\cite{shao2022detecting}, and the results show a consistent trend, indicating explanations also help generalization ability.

\begin{table}[t]
    \begin{subtable}[t]{0.5\textwidth}
    \begin{adjustbox}{width=0.95\linewidth,center}
    \begin{tabular}{r c c c c c}
    \hline
   Dataset & Method & Acc$\uparrow$ & Recall$\uparrow$ & Precision$\uparrow$ & F1$\uparrow$\\
        \hline
\multirow{2}{*}{DD-VQA} & Det. &  $0.7978$ & $\mathbf{0.9639}$ & $0.7588$ & $0.8492$ \\
       &  Det.+exp. & $\mathbf{0.8168}$ & $0.9596$ & $\mathbf{0.7861}$ & $\mathbf{0.8642}$  \\
       \hline
     \multirow{2}{*}{SeqDeepfake~\cite{shao2022detecting}}
        &  Det. & $0.5216$ & $\mathbf{0.9739}$ & $0.5156$ & $0.6742$ \\
        & Det.+exp. & $\mathbf{0.5648}$ & $0.9513$ & $\mathbf{0.5521}$ & $\mathbf{0.6987}$ \\
       \hline
    \end{tabular}
    \end{adjustbox}
    \end{subtable}
    \begin{subtable}[t]{0.5\textwidth}
   \begin{adjustbox}{width=0.95\linewidth,left}
    \begin{tabular}{r c c c c}
    \hline
    Method & Acc$\uparrow$ & Recall$\uparrow$ & Precision$\uparrow$ & F1$\uparrow$\\
        \hline
        Efficient ViT~\cite{coccomini2022combining} &  $0.6849$ & $0.7259$ & $0.7538$ & $0.7396$ \\
        Conv. Cross ViT~\cite{coccomini2022combining}& $0.7763$ & $0.7778$ & $0.8468$ & $0.8108$ \\
        BLIP-TI (ours) & $\mathbf{0.8719}$ & $\mathbf{0.9367}$ & $\mathbf{0.8757}$ & $\mathbf{0.9052}$ \\
      \hline 
    \end{tabular}
    \end{adjustbox}
    \end{subtable}
    \caption{Ablation Studies. LEFT: BLIP trained on datasets w/o explanations. Det.: Detection, exp.: explanations. RIGHT: ViT-based Deepfake Detection Models.}
    \label{tab:ablation}
\end{table}


\Paragraph{Detection performance on fine-grained questions.} In Tab.~\ref{tab:ablation fine}~(right), we analyze the model's performance across different fine-grained questions, and the results indicate that the model consistently achieves satisfactory detection results for all specific questions. We hypothesize that the model's higher performance on skin-related questions can be attributed to the relatively higher number of question-answer pairs related to skin in our dataset, as shown in Fig.~\ref{fig:DD-VQA Dataset}~(b)(2).

\Paragraph{Comparison with ViT-based deepfake detection models.} 
As our model is a multi-modal Transformer, we compare it with pure ViT Transformer-based deepfake detection models to assess whether adding additional textual modality contributes to performance improvement. We compare against Efficient ViT~\cite{coccomini2022combining} and Convolutional Cross ViT~\cite{coccomini2022combining}. 
We evaluate their trained model with images in the DD-VQA dataset and compare them with our answers to general questions of assessing the authenticity of the entire image. The results have been shown in Tab.\ref{tab:ablation}~(right). While Convolutional Cross ViT can perform much better on Efficient ViT, their results are still significantly below our model, even though both models have been trained on the FF++ dataset.



\SubSection{Qualitative Study}
\Paragraph{Qualitative Examples.} We provide qualitative examples in Fig.~\ref{fig:qualitative example}.
We show answers generated by Mini-GPT4~\cite{zhu2023minigpt}, one of the powerful VL pre-trained models, and compare them with our answers for the same question and image.  Mini-GPT4 tends to perceive every image as real, offering detailed descriptions of human facial components, clothing, and the background.
In contrast, our DD-VQA excels at providing better detection results and accurately explaining the reasons behind an image's authenticity or fakeness.

\begin{figure}
    \centering
\includegraphics[width=0.95\linewidth, height=1.2in]{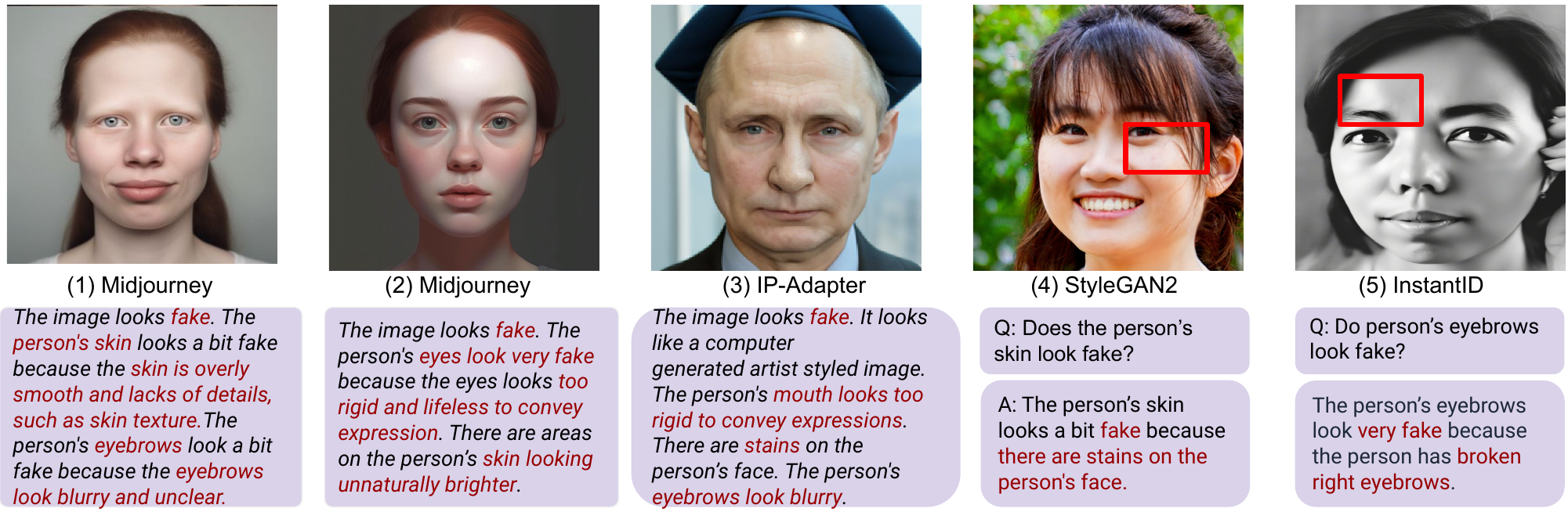}
    \caption{Evaluation of Deepfake Images Beyond the FF++ Dataset.}
    \label{fig:mid-journy}
\end{figure}

\Paragraph{Heatmap Visualization.} In Tab.~\ref{tab:ablation fine}~(left), we illustrate an example to visualize the attention heatmap of the last cross-attention layer in the image-grounded encoder. This visualization aims to assess the model's ability to identify specific regions given the fine-grained questions. We experiment with BLIP and BLIP-TI, and the results show that BLIP alone is already capable of roughly identifying the corresponding region, while the incorporation of our contrastive losses enhances the accuracy of localization, such as ``\textit{eyes}'' and ``\textit{mouth}'' in the example.

\Paragraph{Evaluation of Other Deepfake Images.} The images in the DD-VQA dataset are generated from  CNN-based models. To evaluate our model's generalization ability, we assess its performance on forged facial images generated by different methods, including StyleGAN2~\cite{karras2020analyzing} and diffusion-based models such as Midjourney\footnote{\url{https://www.midjourney.com/}}, IP-Adapter-Face~\cite{ye2023ip-adapter}, and InstantID~\cite{wang2024instantid}. As shown in Fig.~\ref{fig:mid-journy}, our model generates reasonable answers to given questions, demonstrating that our method is not limited to conventional FF++ and can be effective for the latest deepfakes. We provide more examples in the Supplementary.


\section{Conclusion}
\label{sec:conclusion}
This paper proposes a novel task that extends deepfake detection from a conventional binary classification to a VQA task. 
We provide an annotated dataset and benchmark for this task. We explore enhancing the baseline model through contrastive learning and improving the current deepfake detector with our multi-modal representations trained on the DD-VQA dataset.
Our experiments demonstrate that the inclusion of textual explanations is beneficial for both detection performance, generalization ability, and interpretability of deepfake detection.

\noindent \textbf{Limitations.} We acknowledge the following limitations: (1) Our method may generate less effective answers when applied to high-quality deepfake images designed to deceive humans. (2) Our dataset needs more exploration of other deepfake images, such as CGI and diffusion-based images. (3) We have yet to explore the performance with other VL backbone models.


\noindent \textbf{Ethic Statement.} We use publicly available datasets FF++, and discourage any use of our methods for malicious purposes. We commit not to conduct experiments on any images depicting medical deformities.

\newpage
\section*{Supplementary}
\renewcommand{\thesection}{\Alph{section}}
\setcounter{section}{0}

In this supplementary material, we provide:

$\diamond$ Additional details about the DD-VQA Data Annotations Scheme

$\diamond$ Additional implementation details about DD-VQA Enhanced Deepfake Detection.

$\diamond$ Additional DD-VQA experimental details including setup and visualizations. 

\Section{DD-VQA Dataset Annotations}

\Paragraph{Annotation Tools.}
Annotations for DD-VQA are collected entirely by crowd workers from Amazon Mechanical Turk (AMT)~\footnote{\url{https://www.mturk.com/}}. The dataset is collected over the course of $3$ months and $3$ iterations of updating annotation schemes. 
Approximate $9000$  Human Intelligence Tasks (HITs) are launched on AMT, where each HIT involves $3$-$6$ questions, answers, and the corresponding images. Each HIT was designed such that workers manage to earn anywhere between $\$6$-$\$8$ per hour, which follows ethical research standards on AMT~\cite{salehi2015we}.

\Paragraph{Fakeness Annotations.}
From Tab.~\ref{tbl:eyebrowf}-Tab.~\ref{tbl:generalf}, we present examples of fine-grained fake facial features and the corresponding descriptions in our dataset. We provide the annotators with fine-grained feature options and use templates to comprise the description with our templates. Some fakenesses require the annotators to provide the corresponding area, for example, ``\textit{left or right eyebrows}. Also, for the question of which area looks unnatural brighter/darer, the answers need to include the corresponding facial areas, such as "\textit{left/right cheeks"}, \textit{"beside the left/right eyes"}, \textit{"around nose"}, etc.

\begin{figure}
    \centering
    \includegraphics[width=0.85\linewidth]{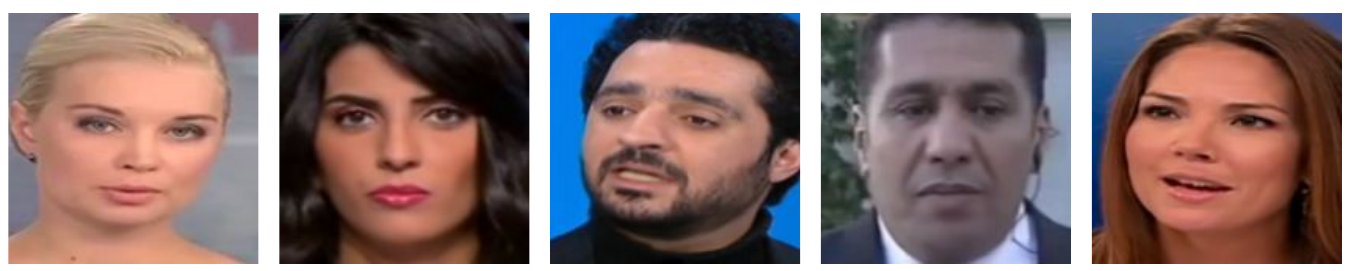}
    \caption{Challenging cases where annotators provide incorrect labels.}
    \label{fig:challenging}
    \vspace{-5mm}
\end{figure}

\Paragraph{Challenging Annotation Cases.} In Fig~\ref{fig:challenging}, we provide examples where at least two annotators mistakenly perceive manipulated images as real. Such cases are excluded when annotators provide inaccurate labels, as effective deception of humans requires the human face in the image to adhere to common-sense knowledge.

\Paragraph{Uncertainty of Fakeness.} There are cases where annotators express uncertainty regarding the image's authenticity. To capture this ambiguity, we offer annotators a fakeness rating scale ranging from $0$ to $5$,  where $0$ and $1$ indicate authenticity, $2$ and $3$ means a slight degree of fakeness, and $4$ and $5$ represent a high degree of fakeness. The corresponding descriptions are ``real'', ``a bit fake'', and ``very fake''.  Annotating the uncertainty of fakeness helps the model simulate human perception of fakeness, thereby enhancing its ability to generate explanations that align more accurately with human judgment.
\begin{figure}
    \centering
    \includegraphics[width=0.85\linewidth]{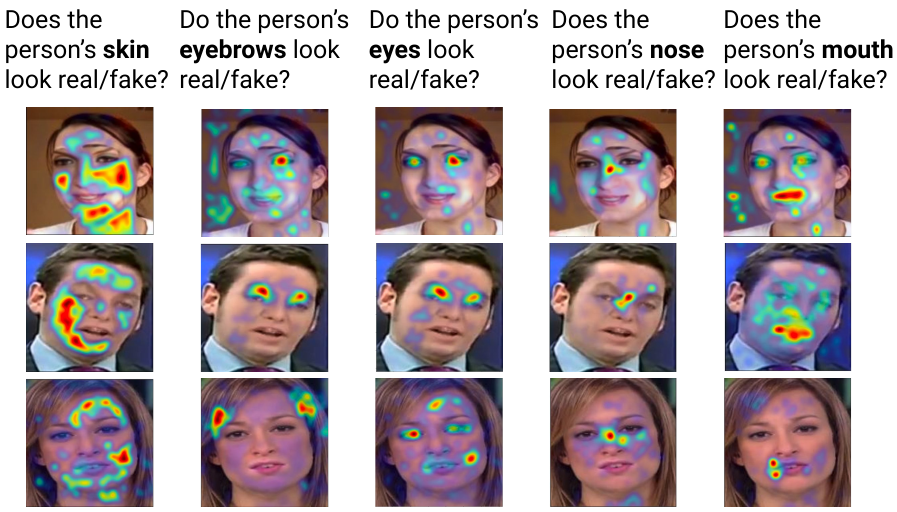}
    \caption{Additional attention heatmap visualization of BLIP-TI.}
    \label{fig:extra visulaiztion}
    \vspace{-3mm}
\end{figure}

\Paragraph{General Questions} assess the overall authenticity of an image. The format of the general question is ``\textit{Does the person in the image look fake?}''. The answers to this question cover the general reasons for authenticity or fakeness. Specifically, the general fakeness factors include ``\textit{obvious manipulated region}'', ``\textit{incomplete face feature}'', ``\textit{unrealistic texture or lighting}'', etc. Conversely, the general reasons for authenticity involve ``\textit{complete face features}'', ``\textit{face features in good shape, size, and positioning.}'', ``\textit{natural expression}'', etc.

\Paragraph{Fine-Grained Facial Feature Questions}  assess the authenticity of individual facial features. There are instances where specific facial components may still exhibit authenticity despite the overall image appearing fake.
The detailed facial features include \textit{eyebrows}, \textit{skin}, \textit{eyes}, \textit{nose}, and \textit{mouth}. The format of the fine-grained feature question is ``\textit{Do the person's X look real/fake?}'', and \textit{X} is any facial component.  We show the corresponding examples in Fig.~\ref{fig:DD-VQA Dataset}.

\begin{figure}[htbp]
    \centering
\includegraphics[width=1.0\linewidth]{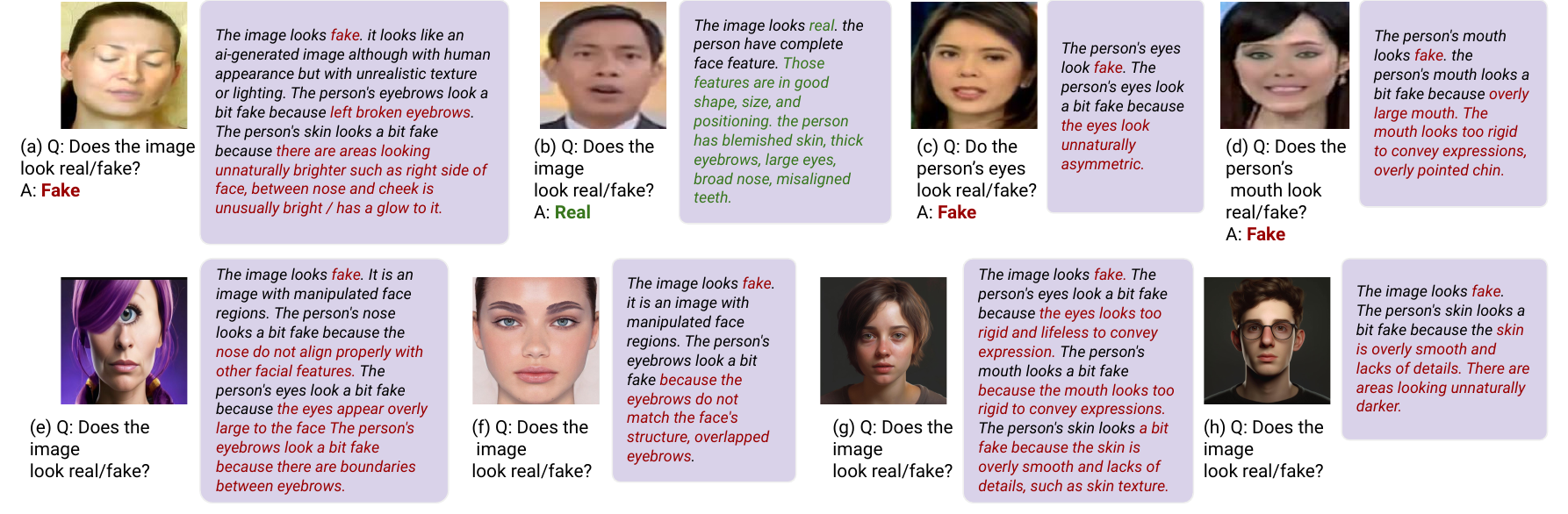}
    \caption{\textbf{Additional Qualitative Examples} (a)-(d) are images from FF++. (e) is a cartoon image; (f) is a Photoshop image showing overlapped eyebrows. (g) and (h) are images from Midjourney.}
    \label{fig:additional qualitative example}
\end{figure}

\begin{itemize}
      \item \textbf{Eyebrows.} Humans commonly have a pair of eyebrows with a symmetrical shape, smooth hair, and a dark color. The presence of overlapping, broken and blurred eyebrows can indicate manipulated images.
    \item \textbf{Skin.} There is no universally ``perfect'' type of skin; however, generally, common skin should exhibit clarity, an even skin tone, and a smooth texture, especially at lower resolutions. Also, the presence of boundaries, discolored patches, or drastically inconsistent skin color on the face are not characteristic of a real person's face.
    \item \textbf{Eyes.} Common eyes include the characteristics of symmetry, clarity, expressiveness, an appropriate size, etc. The blurred and asymmetric eyes in the manipulated image can indicate fakeness.
    \item \textbf{Nose.} An ideal nose should be appropriately positioned, with clear and proportionate nostrils in terms of shape and size. However, the unnaturally curved nose or nose without fine lines are obvious fake signs. 
    \item \textbf{Mouth.} The mouth in our annotation scheme refers to mouth areas, including lips, teeth and chin. The appearance of inappropriate size and color of these areas could be used to indicate fakeness.
\end{itemize}

\vspace{-5mm}
\Section{DD-VQA Enhanced Deepfake Detection}
Our proposed DD-VQA generates multi-modal representations that can serve as a model-agnostic enhancement for general binary deepfake detectors.  We illustrate our approach using RECCE~\cite{cao2022end} as an example.
RECCE proposes a forgery detection framework that leverages the common compact representations of genuine faces based on reconstruction classification learning. 
Specifically, the images are fed into an encoder-decoder reconstruction network for representation learning. The encoder's output, denoted as $\mathbf{F}_{1}$, undergoes a multi-scale graph reasoning module to enhance better representation, denoted as $\mathbf{F}_{2}$, which is subsequently combined with $\mathbf{F}_{1}$.
In summary, the vision representation of deepfake detection is $\mathbf{F}^{\prime}=\mathbf{F}_{1}+\mathbf{F}_{2}$.
Based on this, we incorporate our DD-VQA enhanced multi-modal representation $\mathbf{F}$ obtained from our VL model trained using the DD-VQA dataset. 
We first utilize a few CNN layers to transform $\mathbf{F}$ into the same shape as $\mathbf{F}^{\prime}$. We can obtain the final enhanced representation $\mathbf{F}^{en.}$ with $\mathbf{F}^{en.} = \mathbf{F}^{\prime} + \theta(\mathbf{F})$, where $\theta(*)$ represents the necessary tensor shape transformations for fusing $\mathbf{F}$ and $\mathbf{F}^{\prime}$.


\begin{table}[t]
    \centering
\footnotesize
        \centering
        \resizebox{1\textwidth}{!}{
\begin{tabular}{rcccc}
\hline
 Method & \multicolumn{2}{c}{DD-VQA Deepfake Detection} & \multicolumn{2}{c}{DD-VQA Answer Generation}  \\
   & Acc $\uparrow$ &F1 $\uparrow$ & BLUE-4 & Rouge\_L\\
   \hline
    No Distortion & $\mathbf{0.8749}$ & $\mathbf{0.9007}$  & $\mathbf{0.4075}$ & $\mathbf{0.6085}$ \\
    Resize(0.75X) & $0.8621$ & $0.8987$ & $0.3987$ & $0.6025$ \\
    JPEGCompression(quality=$75$) & $0.8593$ & $0.8827$ & $0.3864$ & $0.5846$ \\
    GaussianNoise($\sigma=3$) & $0.8434$ & $0.8676$ & $0.3813$ & $0.5811$\\
    Color Enhancement(factor=3.0) & $0.8385$ & $ 0.8639$ & $0.3792$ & $0.5761$\\
\hline
\end{tabular}
}
\caption{Robustness Evaluation.}
\label{tab:robust evaluation}
\end{table}

\Section{Experiment Setup}
\Paragraph{Metrics} We mainly use image-caption-based metrics to evaluate the quality of the generated text, as follows.
\begin{itemize}
      \item \textbf{BLEU-4}~\cite{papineni2002bleu} is used to evaluate the precision of the match between the generated text and reference text based on 4-grams.
      \item \textbf{CIDEr}~\cite{vedantam2015cider} measures the consensus between the generated text and the referenced text, considering both word and grammar similarity and the alignment in terms of meaning and content.
      \item \textbf{Rouge\_L}~\cite{lin2004rouge} evaluates the \textbf{L}ongest \textbf{C}ommon \textbf{S}ubsequence (LCS) of words between the generated text and the referenced text. Using LCS does not require consecutive matches but in-sequence matches reflecting sentence-level word order.
      \item  \textbf{METEOR}~\cite{denkowski2014meteor} considers precision, recall, stemming, synonymy, and word order. It employs a unigram-based matching approach but extends it with additional semantic features.
      \item \textbf{SPICE}~\cite{anderson2016spice} evaluates how well a generated text can capture the specific entities present in the image, emphasizing precision, recall, and diversity.
\end{itemize}

\Paragraph{ViT-based deepfake detection models.} Efficient ViT combines a ViT with a convolutional EfficientNet B0 as the feature extractor. Convolutional Cross ViT builds upon both the Efficient ViT and the multi-scale Transformer, and enable the utilization of larger patches to achieve a broader receptive field. Although both Efficient ViT and Convolutional Cross ViT use video deepfake datasets (FF++~\cite{rossler2019faceforensics++} and DFDC~\cite{dolhansky2020deepfake}), they extract frames from videos and use images for model training.

\begin{table}[h!]
  \centering
  \begin{tabular}{   m{1.7cm} | m{3cm} | m{6cm}  }
    \toprule[2pt]
    Fine-grained Features & Images & Descriptions \\ 
    \toprule[2pt]
    Overlapped eyebrows &   
    \begin{minipage}{.3\textwidth}
    \includegraphics[width=25mm, height=25mm]{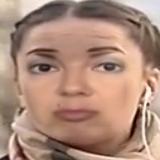}
    \end{minipage} &   The person's eyebrows look fake. The person's eyebrows look very fake because the person has left overlapped eyebrows and right-overlapped eyebrows. \\
    Broken eyebrows &   \begin{minipage}{.3\textwidth}
      \includegraphics[width=25mm, height=25mm]{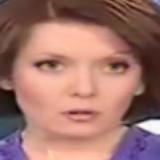}
    \end{minipage} & The person's eyebrows look fake. The person's eyebrows look very fake because the person has broken left eyebrow. \\
    Blurry eyebrows
    &    
    \begin{minipage}[t][2cm][t]{.3\textwidth}
      \includegraphics[width=25mm, height=25mm]{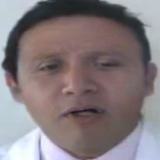}
    \end{minipage}
    & The person's eyebrows look fake. The person's eyebrows look very fake because the eyebrows look blurry and unclear. \\
    Boundary between eyebrows&   \begin{minipage}{.3\textwidth}
      \includegraphics[width=25mm, height=25mm]{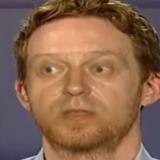}
    \end{minipage} & The person's eyebrows look fake. The person's eyebrows look fake because there is a boundary between the person's eyebrows.  \\
    \bottomrule[2pt]
  \end{tabular}
  \caption{Fake Eyebrows Features.}\label{tbl:eyebrowf}
\end{table}

\begin{table}[h!]
  \centering
  \begin{tabular}{  m{1.7cm} | m{3cm} | m{6cm}  }
    \toprule[2pt]
    Fine-grained Features & Images & Descriptions \\
    \toprule[2pt]
    Blurry eyes &    
    \begin{minipage}[t][2cm][t]{.3\textwidth}
      \includegraphics[width=25mm, height=25mm]{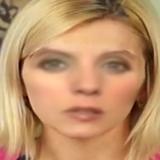}
    \end{minipage}
    &  The person's eyes look fake. The person's eyes look fake because the eyes look blurry and unclear.\\
    Unnatural asymmetric eyes & 
    \begin{minipage}[t][2cm][t]{.3\textwidth}
      \includegraphics[width=25mm, height=25mm]{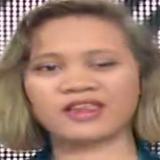}
    \end{minipage}
    & The person's eyes look fake. The person's eyes look fake because the person has unnatural asymmetric eyes.
    \\
    Rigid Eyes & 
     \begin{minipage}[t][2cm][t]{.3\textwidth}
      \includegraphics[width=25mm, height=25mm]{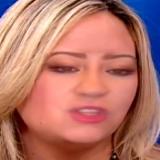}
    \end{minipage} & The person's eyes look fake. The person's eyes look fake because the person's eyes are too rigid to convey expressions.
    \\ 
    \bottomrule[2pt]
  \end{tabular}
  \caption{Fake Eyes Features.}\label{tbl:eyes}
\end{table}

\begin{table}[h!]
  \centering
  \begin{tabular}{  m{1.7cm} | m{3cm} | m{6cm}  }
    \toprule[2pt]
    Fine-grained Features & Images & Descriptions \\
    \toprule[2pt]
    Boundaries
    &    
    \begin{minipage}[t][2cm][t]{.3\textwidth}
      \includegraphics[width=25mm, height=25mm]{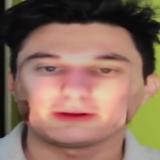}
    \end{minipage}
    &  The person's skin looks fake. The person's skin looks very fake because there are boundaries on the person's face, such as boundaries on the person's left and right cheeks. \\
    Inconsistent skin color  &   
    \begin{minipage}{.3\textwidth}
      \includegraphics[width=25mm, height=25mm]{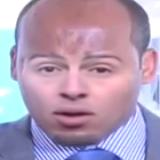}
    \end{minipage} & The person's skin looks fake. The person's skin looks very fake because the person has inconsistent skin color. \\
    Discolored patches &  \begin{minipage}{.3\textwidth}
      \includegraphics[width=25mm, height=25mm]{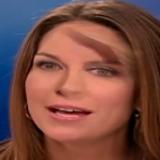}
    \end{minipage} & The person's skin looks fake. The person's skin looks very fake because there is a discolored path on the person's forehead.
    \\ 
    \bottomrule[2pt]
  \end{tabular}
  \caption{Fake Skin Features.}\label{tbl:skinf}
\end{table}

\begin{table}[h!]
  \centering
  \begin{tabular}{  m{1.7cm} | m{3cm} | m{6cm}  }
  \toprule[2pt]
    Fine-grained Features & Images & Descriptions \\ 
    \toprule[2pt]
    Unnaturally curved nose &    
    \begin{minipage}[t][2cm][t]{.3\textwidth}
      \includegraphics[width=25mm, height=25mm]{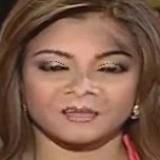}
    \end{minipage}
    &  The person's nose looks fake. The person's nose looks unnaturally curved. \\
    nose lacks of details & 
     \begin{minipage}[t][2cm][t]{.3\textwidth}
      \includegraphics[width=25mm, height=25mm]{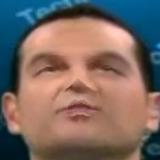}
    \end{minipage}
    &  The person's nose looks fake. The person's nose looks very fake because the nose lacks of pores and fine lines.  \\
    \bottomrule[2pt]
  \end{tabular}
  \caption{Fake Nose Features.}\label{tbl:nosef}
\end{table}

\begin{table}[h!]
  \centering
  \begin{tabular}{  m{1.7cm} | m{3cm} | m{6cm}  }
    \toprule[2pt]
    Fine-grained Features & Images & Descriptions \\
    \toprule[2pt]
    Blurry Mouth &    
    \begin{minipage}[t][2cm][t]{.3\textwidth}
      \includegraphics[width=25mm, height=25mm]{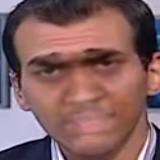}
    \end{minipage}
    &  The person's mouth area looks fake. The person's mouth looks blurry and unclear. \\
    Mouth with unnatural color &   
    \begin{minipage}[t][2cm][t]{.3\textwidth}
      \includegraphics[width=25mm, height=25mm]{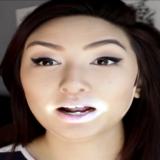}
    \end{minipage} &  The person's mouth area looks fake. The person's mouth shows an unnatural white color.\\
    unnatural coloring/blurry teeth  &  \begin{minipage}[t][2cm][t]{.3\textwidth}
      \includegraphics[width=25mm, height=25mm]{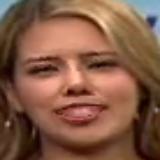}
    \end{minipage} &   The person's mouth area looks fake. The person's teeth look misaligned with the rest of the mouth. The person's teeth look unnatural coloring.\\
    \bottomrule[2pt]
  \end{tabular}
  \caption{Fake Mouth Features.}\label{tbl:mouthf}
\end{table}

\begin{table}[h]
  \centering
  \begin{tabular}{  m{1.7cm} | m{3cm} | m{6cm}  }
      \toprule[2pt]
    Fine-grained Features & Images & Descriptions \\
       \toprule[2pt]
    Incomplete facial features &    
    \begin{minipage}[t][2cm][t]{.3\textwidth}
      \includegraphics[width=25mm, height=25mm]{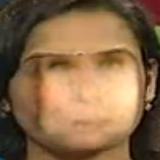}
    \end{minipage}
    &  The image looks fake because the person has incomplete facial features. \\
     Unclear eyeglass frame&    
    \begin{minipage}[t][2cm][t]{.3\textwidth}
      \includegraphics[width=25mm, height=25mm]{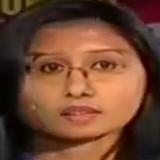}
    \end{minipage}
    &  The image looks fake because the person's eyeglass frame looks unclear.\\
    Mustache &    
    \begin{minipage}[t][2cm][t]{.3\textwidth}
      \includegraphics[width=25mm, height=25mm]{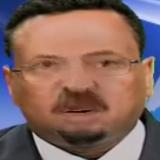}
    \end{minipage}
    &  The image looks fake because the person's mustache does not align with other facial features. \\
   \bottomrule[2pt]
  \end{tabular}
  \caption{General Fake Features.}\label{tbl:generalf}
\end{table}


\Section{Qualitative Study}
\Paragraph{Visualization.} We present additional visualization examples in Fig.\ref{fig:extra visulaiztion} generated by our best model BLIP-TI. The model is trained with both language modeling loss and our designed contrastive losses. These examples demonstrate that the highlighted attention areas predominantly align with the facial components mentioned in the question. We employ GradCam~\cite{selvaraju2017grad} visualization technique to show the alignments between textual tokens and the highlighted area in the image.

\Paragraph{Robustness Evaluation.} We conduct a robustness evaluation of our model, considering aspects such as resizing, compression, Gaussian noise, and color enhancement. We evaluate both detection and explanation generation performances. As shown in Tab.~\ref{tab:robust evaluation}, our model appears to be robust to different variations, especially regarding the quality of generated textual explanations.

\Paragraph{Qualitative Examples} We provide additional qualitative examples in Fig.~\ref{fig:additional qualitative example}. We extend our testing beyond the FF++ dataset. We evaluate our model on diverse images, including cartoon images, Photoshop images, and images generated using a diffusion model. 
These examples show our model can capture common-sense knowledge of human facial features well. For instance, the cartoon image of Fig.~\ref{fig:additional qualitative example}~(e), our model can capture the pattern of ``\textit{over large eyes}''. Also, we manipulate a real image to put another pair of eyebrows on top of the original eyebrows, as shown in Fig.~\ref{fig:additional qualitative example}~(f), and our model still can capture the fakeness of  ``\textit{overlapped eyebrows}''. For images from Midjourney (Fig.~\ref{fig:additional qualitative example}~(g) and~(h)), our model can capture the fakeness of ``\textit{rigid eyes and mouth}''.

\clearpage  

%
%
\bibliographystyle{splncs04}
\bibliography{main}
\end{document}